\documentclass[10pt,twocolumn,letterpaper]{article}

\usepackage[pagenumbers]{cvpr} 

\usepackage{graphicx}
\usepackage{amsmath}
\usepackage{amssymb}
\usepackage{booktabs}

\usepackage{booktabs}
\usepackage{multirow}
\usepackage{nicefrac}      
\usepackage{algorithm}
\usepackage{algpseudocode}
\usepackage{caption}
\usepackage{bm}

\usepackage{colortbl}

\usepackage{amsmath,amssymb,amsfonts}

\definecolor{Gray}{gray}{0.92}

\definecolor{cvprblue}{rgb}{0.21,0.49,0.74}
\usepackage[pagebackref,breaklinks,colorlinks,allcolors=cvprblue]{hyperref}

\def\paperID{11771} 
\def\confName{CVPR}
\def\confYear{2025}

\title{Improving Transferable Targeted Attacks with Feature Tuning Mixup}

\author{Kaisheng Liang \qquad Xuelong Dai \qquad Yanjie Li \qquad Dong Wang \qquad Bin Xiao\thanks{Corresponding author.}\\
Department of Computing, The Hong Kong Polytechnic University\\
{\tt\small cskliang@comp.polyu.edu.hk \qquad b.xiao@polyu.edu.hk}\\
{\tt\small \{xuelong.dai, yanjie.li, dong-comp.wang\}@connect.polyu.hk}\\
}

\begin{document}
\maketitle
\begin{abstract}
Deep neural networks (DNNs) exhibit vulnerability to adversarial examples that can transfer across different DNN models. A particularly challenging problem is developing transferable targeted attacks that can mislead DNN models into predicting specific target classes. While various methods have been proposed to enhance attack transferability, they often incur substantial computational costs while yielding limited improvements. Recent clean feature mixup methods use random clean features to perturb the feature space but lack optimization for disrupting adversarial examples, overlooking the advantages of attack-specific perturbations. In this paper, we propose Feature Tuning Mixup (FTM), a novel method that enhances targeted attack transferability by combining both random and optimized noises in the feature space. FTM introduces learnable feature perturbations and employs an efficient stochastic update strategy for optimization. These learnable perturbations facilitate the generation of more robust adversarial examples with improved transferability. We further demonstrate that attack performance can be enhanced through an ensemble of multiple FTM-perturbed surrogate models. Extensive experiments on the ImageNet-compatible dataset across various DNN models demonstrate that our method achieves significant improvements over state-of-the-art methods while maintaining low computational cost.
\footnote{Our code is available at https://github.com/uhiu/feature-tuning-mixup}
\end{abstract}
    
\section{Introduction}
\label{sec:introduction}

Deep neural networks (DNNs) are highly vulnerable to adversarial attacks~\cite{GoodfellowSS14,KurakinGB17a,MadryMSTV18}, raising critical security concerns.
In image classification tasks, adversarial attacks can deceive DNNs by adding imperceptible perturbations to benign images, causing the models to produce erroneous predictions.
More concerningly, research~\cite{DongLPS0HL18,XieZZBWRY19,DongPSZ19} has demonstrated that adversarial examples possess transferability properties: adversarial examples generated using one white-box model can successfully fool other black-box models without requiring access to their architectures or parameters. 
This cross-model transferability of adversarial examples poses severe security threats to real-world AI systems.
Therefore, studying the transferability of adversarial examples is crucial for enhancing the robustness of DNNs against such security vulnerabilities.

\begin{figure}[t]
   \centering
    \includegraphics[width=0.95\linewidth]{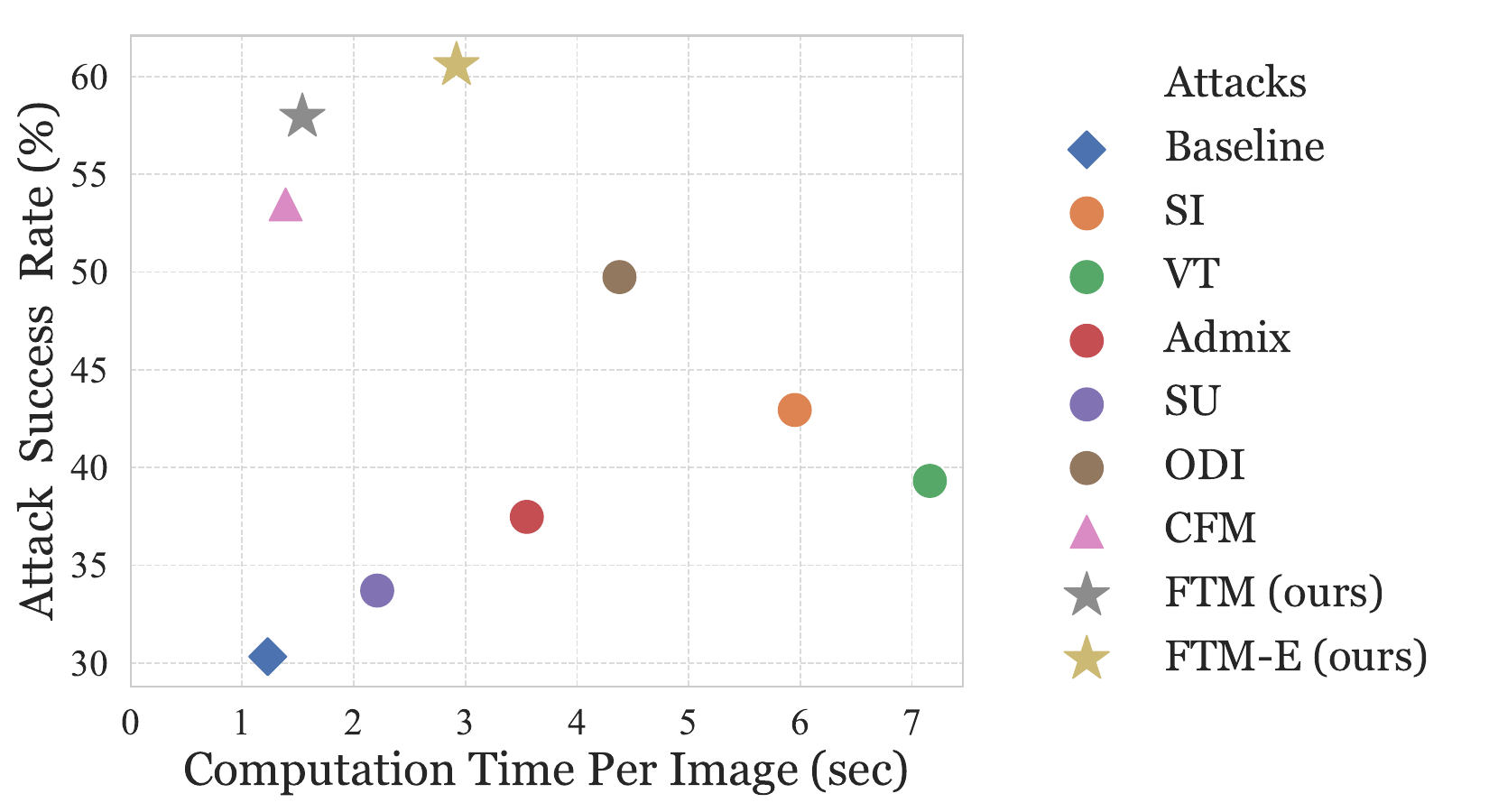}
 
    \caption{%
      \textbf{Effectiveness and efficiency of targeted attacks.} Average attack success rates on 14 black-box models, along with the computation time required to generate an adversarial example. Our methods demonstrate superior performance with low computational cost, surpassing state-of-the-art methods.
    }
    \label{fig:starter}
    \vspace{-0.3cm}
\end{figure}

 \begin{figure*}[!ht]
    \centering
     \includegraphics[width=0.91\linewidth]{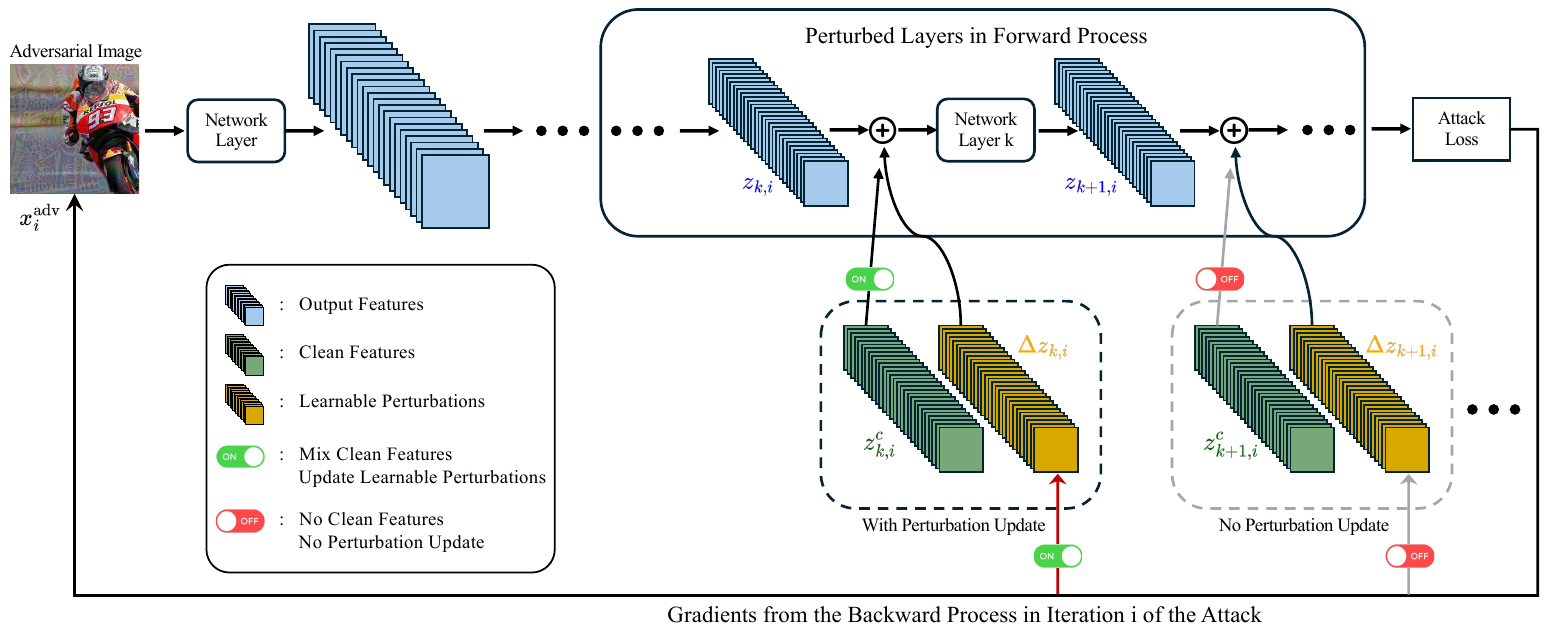}
  
     \caption{
       Overview of our FTM attack.
       In the forward pass, the learnable perturbations are added to the output features.
       Only a small portion of network layers are randomly selected for perturbation updates and clean feature mixup in each attack iteration.
       In the backward pass, the gradients of the selected perturbations and adversarial image are computed together for update.
     }
     \label{fig:overview}
     \vspace{-0.2cm}
 \end{figure*}

Targeted transfer-based attacks aim to mislead models into predicting specific target classes, which remains a significant challenge. Due to the lack of knowledge about the target model's training data and architecture, such attacks tend to generate targeted adversarial examples that overfit to surrogate models and fail to transfer to target models. 
Various transferable targeted attacks~\cite{Byun_2022_CVPR,Zhao_2021_NeurIPS,Wei_2023_CVPR} have been proposed but still leave much room for improvement.
Notably, data augmentation methods~\cite{Wang021,LinS00H20,Wang_2021_admix} are commonly employed for targeted attacks.
But they primarily operate in the image space, while augmentation methods at the feature level remain relatively unexplored.
Recently, the Clean Feature Mixup (CFM) method~\cite{byun2023introducing} has made important progress by using clean features from random images to disrupt adversarial examples, leading to better targeted attack success rates (Figure~\ref{fig:starter}).
However, CFM uses random clean features without optimizing them specifically for disrupting adversarial examples.
We hypothesize that carefully optimized perturbations would be more effective in disrupting adversarial examples, potentially yielding substantial improvements in attack transferability.

In this work, we propose Feature Tuning Mixup (FTM), a novel attack method that optimizes attack-specific feature perturbations to enhance the transferability of targeted adversarial examples.
Our method fine-tunes the intermediate features of surrogate models during the attack process by introducing learnable feature perturbations.
These feature perturbations are optimized by mixing them with clean features from random images to counteract the adversarial effects of targeted adversarial examples.
Through this mechanism, adversarial examples are gradually compelled to become more robust, allowing them to overcome these feature-level disturbances.
To address the computational overhead of optimizing feature perturbations, we propose a stochastic update strategy.
This strategy initializes the learnable perturbations using values from previous attack iterations and selectively optimizes only a small subset of feature perturbations in each iteration.
Ultimately, FTM significantly improves the transferability of targeted adversarial examples while maintaining computational efficiency.

We provide an overview of the proposed FTM in Figure~\ref{fig:overview}. FTM first introduces learnable perturbations into the deep layers of the surrogate model.
The learnable perturbations are added element-wise to the original output features before passing to the next layer. In each attack iteration, a small subset of layers is randomly selected and combined with random clean features.
Finally, we compute gradients for the learnable perturbations and adversarial image using only one forward and backward pass, similar to standard attacks. 
We update the selected learnable perturbations and adversarial image, while unselected perturbations remain unchanged.
As a result, FTM fully leverages optimized attack-specific feature perturbations, prevents overfitting to surrogate models, and significantly enhances the transferability of targeted attacks.

Our contributions are summarized as follows:

\begin{itemize}
   \item We revisit feature-level augmentation methods for transferable targeted attacks and identify that incorporating attack-specific feature perturbations, previously overlooked, can efficiently boost existing attacks.
   
   \item We introduce Feature Tuning Mixup (FTM), a novel approach that combines random and optimized feature perturbations to enhance targeted transfer-based attacks. Our method includes an efficient stochastic update strategy that maintains computational efficiency while improving attack transferability.
   
   \item We conduct extensive experiments across various source and target DNN models on the ImageNet-compatible dataset, demonstrating that our FTM achieves a significant improvement over state-of-the-art methods.
\end{itemize}

\section{Related Work}
\label{sec:related}
Adversarial attacks~\cite{carlini2017towards,eykholt2018robust,IlyasSTETM19,GoodfellowSS14,MadryMSTV18} have been widely studied in the field of adversarial learning, including unrestricted attacks~\cite{dai2025advdiff}, physical attacks~\cite{li2023physical,li2025uvattack}, and black-box attacks~\cite{BrendelRB18_2018_iclr,liang2023styless,DongLPS0HL18}. 
Notably, black-box attacks can be categorized into query-based~\cite{BrendelRB18_2018_iclr} and transfer-based attacks~\cite{DongLPS0HL18,liang2023styless}. 
The latter, which does not require access to the target DNN model's outputs, parameters or architecture, is more practical and is the focus of this paper.

\subsection{Non-targeted Transfer-based Attacks}

Current non-targeted transfer attacks are primarily based on the Iterative-Fast Gradient Sign Method (I-FGSM)~\cite{KurakinGB17a}. This approach iteratively computes the gradient of the classification loss with respect to the input and updates adversarial examples in the direction that maximizes the loss. However, it often leads to overfitting on white-box models, making the adversarial examples less effective against other black-box models. Numerous studies have explored solutions to the overfitting issue to enhance adversarial transferability. The Momentum Iterative FGSM (MI-FGSM)~\cite{DongLPS0HL18} incorporates a momentum-based optimization method during iterative attacks. Translation-Invariant (TI-FGSM) method~\cite{DongPSZ19} enhances the transferability of adversarial examples by achieving translation invariance through gradient smoothing. Data augmentation techniques are widely used to mitigate overfitting. The Diverse Inputs (DI) method~\cite{XieZZBWRY19} employs random resizing and padding with a probability $p$ to generate diverse input, while the Resized Diverse Inputs (RDI) method~\cite{zou2020improving} fixes $p=1$ and rescales DI-enhanced images back to their original size. The Scale-Invariant (SI) method~\cite{LinS00H20} generates adversarial perturbations using multiple scale transformations, and Admix~\cite{Wang_2021_admix} extends SI by blending images from other labels into the input. Variance Tuning (VT) method~\cite{Wang021} randomly selects data points around a given data point and computes their gradients to avoid overfitting. These methods operate in the image space. Feature-level approaches have also been studied to improve adversarial transferability. For instance, the Feature Importance-aware Attack (FIA)~\cite{Wang_2021_ICCV} identifies and perturbs important features in the intermediate layers of a surrogate model. The Skip Gradient Method (SGM)~\cite{Wu0X0M20} utilizes more gradients from skip connections. Ensemble-based attacks~\cite{LiuCLS17,DongLPS0HL18} utilize multiple surrogate models to generate adversarial examples.

\subsection{Targeted Transfer-based Attacks}

Targeted attacks are more challenging than non-targeted attacks because they not only aim to cause incorrect predictions by the models but also to mislead the models into predicting a specific target class. We can categorize targeted attacks into two types based on whether they require a training dataset. Previous works have utilized training datasets to generate auxiliary modules~\cite{InkawhichLCC20,InkawhichLWICC20} for transfer-based attacks or directly train a model to generate adversarial examples~\cite{Naseer_2021_ICCV,yang2022boosting,wang2023towards}. This paper primarily focuses on transfer-based attacks that do not require a training dataset.

Targeted transfer-based attacks are widely studied in adversarial machine learning due to their practicality, as they do not require access to the target model's architecture and parameters, nor do they need training datasets.
The Activation Attack (AA)~\cite{Inkawhich_2019_CVPR} proposes using the distance between adversarial examples and targeted images in feature space as a loss function, but its attack effectiveness may depend on the selection of targeted images.
Po+Trip~\cite{Li_2020_CVPR} employs Poincare distance to adapt the gradients' magnitude and uses triplet loss to address insufficient inter-class distances, but it only works well on target models with similar architectures.
Logit Attack (Logit)~\cite{Zhao_2021_NeurIPS} discovers that a simple logit loss can generate better transferable adversarial examples than CE loss, significantly outperforming previous methods.
Self-Universality (SU) method~\cite{Wei_2023_CVPR} finds that highly universal adversarial perturbations tend to be more transferable for targeted attacks and proposes optimizing the perturbations in different local regions of the image.
The Object-based Diverse Input (ODI) method~\cite{Byun_2022_CVPR} is an efficient data augmentation in image space for targeted attacks using random 3D object and rendering.

Recently, the Clean Feature Mixup (CFM) method~\cite{byun2023introducing} proposes a feature-level augmentation which achieves state-of-the-art performance in targeted transfer-based attacks.
It stores clean features as random feature perturbations, which mitigates the overfitting to surrogate models.
However, the proposed clean features in CFM lack specific optimization for targeted attacks, leaving significant room for improvement.
We develop the Feature Tuning Mixup (FTM) method, which effectively and efficiently optimizes attack-specific feature perturbations and significantly improves transferable targeted attacks.

\textbf{Differences from intermediate-level attacks.}
This paper investigates feature-level augmentation for targeted attacks, which differs from existing intermediate-level attacks~\cite{HuangKGHBL19,LiGC20,Wang_2021_ICCV,zhang2022improving}.
Our FTM is based on the transfer-based attack framework~\cite{KurakinGB17a} with logit loss~\cite{Zhao_2021_NeurIPS}, which is currently the most effective framework for targeted transfer-based attacks.
In contrast, intermediate-level attacks use loss functions based on a specific intermediate layer, which is hard to apply to targeted attacks.
For instance, FIA~\cite{Wang_2021_ICCV} and NAA~\cite{zhang2022improving} require identifying important features based on the true label, while targeted attacks focus on the target label instead of the true label.
ILA~\cite{HuangKGHBL19} requires an existing adversarial example as guidance and cannot function as a standalone targeted attack.
Therefore, we do not compare FTM with intermediate-level attacks in experiments.

\textbf{Comparison with adversarial learning methods.}
Our method learns feature perturbations in an adversarial manner without requiring additional training datasets.
While some existing attacks~\cite{wu2021improving,qin2022boosting} also employ adversarial strategies, they operate in the image space and typically require access to training datasets.
For instance, ATTA~\cite{wu2021improving} improves transferability by training an adversarial transformation network to destroy adversarial examples.
However, this approach demands substantial computational resources and training data.
Like previous work~\cite{Wei_2023_CVPR,Zhao_2021_NeurIPS,Byun_2022_CVPR,byun2023introducing}, we study targeted transfer attacks without extra data and do not compare with methods that need training data.
RAP~\cite{qin2022boosting} utilizes adversarial learning to identify worst-case perturbations in the image space, aiming to locate regions with flat loss landscapes.
However, this process requires numerous additional backpropagation steps, leading to prohibitively high computational costs.
In contrast, we propose a novel approach that optimizes feature perturbations directly in the multi-layer feature space, an unexplored direction in previous research.
Through this innovation, our method achieves superior practicality, effectiveness, and efficiency compared to existing methods.

\section{Methodology}
\label{sec:methodology}

\subsection{Preliminaries}

\textbf{Problem formulation.}
For targeted transfer-based attacks, the attacker is assumed to have a white-box surrogate model $F$, which is a pre-trained image classifier in this paper.
We assume that the attacker has no additional training dataset.
Given a benign image $x$, a surrogate model $F$ and a target label $y_t$, which is different from the true label $y$ of $x$.
The goal of the attack is to generate an adversarial example $x^{\text{adv}}$ that is misclassified as $y_t$ by $F$. The difference between $x$ and $x^{\text{adv}}$ should be subtle, and following previous work, we use the $\ell_{\infty}$-norm constraint.
The optimization objective can be formulated as:
\begin{equation}
    \arg\min_{x^{\text{adv}}} \mathcal{L}(F(x^{\text{adv}}), y_t), \quad s.t. \quad \|x - x^{\text{adv}}\|_{\infty} \leq \epsilon,
    \label{eq:threat_model}
\end{equation}
where $\mathcal{L}$ is the adversarial loss, e.g., the cross-entropy loss or logit-based loss, and $\epsilon$ is the perturbation budget.

\textbf{MI-FGSM.}
To solve the problem above, existing methods usually adopt the iterative gradient-based technique, e.g., the MI-FGSM~\cite{DongLPS0HL18}, formulated as follows:
\begin{align}
    &g_{i+1} = \nabla_x \mathcal{L}(F(x^{\text{adv}}_i), y_t), \label{eq:i_fgsm_grad} \\
    &\tilde g_{i+1} = \mu \cdot g_{i} + g_{i+1}/\| g_{i+1}\|_1, \label{eq:i_fgsm_momentum} \\
    &x^{\text{adv}}_{i+1} = x^{\text{adv}}_i - \eta \cdot \text{sign} \left( \tilde g_{i+1} \right), \label{eq:i_fgsm_update} \\
    &x^{\text{adv}}_{i+1} = \text{Clip}_{x, \epsilon} \left( x^{\text{adv}}_{i+1} \right), \label{eq:i_fgsm_clip}
\end{align}
where $g_i$ is the gradient of the loss function with respect to the input image $x$ in the $i$-th iteration.
$\mu$ is the momentum decay factor, and $\eta$ is the step size.
$x^{\text{adv}}_{i+1}$ is clipped within $\epsilon$ bound.
Initially, we set $x^\text{adv}_0=x$ and $g_0=0$.

\textbf{CFM.}
To improve the transferability of targeted attacks, the Clean Feature Mixup (CFM) method~\cite{byun2023introducing} leverages stored clean features to augment adversarial examples, achieving state-of-the-art performance among existing targeted transfer-based attacks.
Specifically, it selects outputs from certain network layers as perturbation targets and uses output features from different benign images as feature perturbations.
To integrate with MI-FGSM, 
the feature augmentation in CFM can be formulated as follows:
\begin{equation}
    {z}^{\prime}_{k,i} = (1 - \alpha_{k,i}) \odot {z}_{k,i} + \alpha_{k,i} \odot {z}_{k,i}^c,
    \label{eq:cfm_perturb}
\end{equation}
where $\odot$ denotes element-wise multiplication, ${z}_{k,i}$ is the input feature to the $k$-th layer in the $i$-th attack iteration of MI-FGSM, ${z}_{k,i}^c$ is a random stored clean feature, and $\alpha_{k,i}$ is a vector representing channel-wise mixing ratios.
The network layer $k$ is randomly selected from pre-defined candidate layers with probability $p$.

\subsection{Analysis of CFM}
CFM is a powerful \emph{feature-level augmentation} method for targeted attacks.
To understand the effectiveness of CFM, we formulate a generalized optimization step for MI-FGSM based on feature-level augmentation as follows:
\begin{align}
    g^{\prime}_{i+1} &= \nabla_x \mathcal{L} \left( F (x^{\text{adv}}_i ; \Delta z_{i}), y_t \right), \label{eq:cfm_grad} \\
    \Delta z_{i} &= \{\Delta z_{k,i}, k=1, \cdots, n \}, \label{eq:delta_def}
\end{align}
where $F (x^{\text{adv}}_i ; \Delta z_{i})$ denotes the surrogate model $F$ with the perturbation $\Delta z_{i}$ added to the network outputs in the $i$-th attack iteration.
Equation~\eqref{eq:cfm_grad} improves upon Equation~\eqref{eq:i_fgsm_grad} by incorporating feature-level perturbations, which is the key idea of CFM.
Notably, the perturbation $\Delta z_{i}$ represents the collection of perturbations for \emph{all} network layers.
We use $\Delta z_{k,i}$ to denote the perturbation for the input of the $k$-th layer, where $n$ is the maximum layer index of $F$.
Initially, we set $\Delta z_{i} = 0$.
For any layer $d$ that does not require perturbation, we simply set $\Delta z_{d,i} \equiv 0$.

CFM uses clean features to construct $\Delta z_i$, and it claims that the rationale is that clean features can suppress the adversarial effect of $x^\text{adv}_i$~\cite{byun2023introducing}.
We formulate this idea in the following equation:
\begin{equation}
    \mathcal{L} \left( F (x^{\text{adv}}_i ; \Delta z_{i}), y_t \right) > \mathcal{L} \left( F (x^{\text{adv}}_i), y_t \right),
    \label{eq:cfm_idea}
\end{equation}
where $\Delta z_{i}$ is supposed to appropriately disrupt adversarial loss in Equation~\eqref{eq:threat_model} for obtaining robust $x^\text{adv}$.

However, clean features are attack-agnostic in nature, as they are not explicitly optimized for the objective in Equation~\eqref{eq:cfm_idea}.
Although CFM demonstrates promising improvements, we argue that there remains unexplored potential in feature-level augmentation for targeted attacks.
We believe that integrating attack-specific optimization with attack-agnostic clean features can create a more powerful feature-level augmentation strategy.

\subsection{Feature Tuning Mixup (FTM)}
\label{sec:ftm-alg}
We propose Feature Tuning Mixup (FTM), a novel method that uses attack-specific feature perturbations to enhance feature-level augmentation methods.
We formulate the optimization problem for learning our feature perturbations and adversarial examples in each attack iteration as follows:
\begin{equation}
    \min_{x^{\text{adv}}_i} \max_{\Delta z_i} \mathcal{L} \left( F (x^{\text{adv}}_i ; \Delta z_i), y_t \right).
    \label{eq:ftm_obj}
\end{equation}

However, the optimization problem above is challenging to solve from two perspectives: effectiveness and efficiency.
Regarding efficiency, traditional adversarial training~\cite{MadryMSTV18} often requires multiple steps to optimize $\Delta z_i$ in each attack iteration, resulting in substantial additional computational cost.
As for effectiveness, since $\Delta z_{i}$ involves a joint problem across multiple iterations and network layers, inappropriate perturbations can damage the attack effect, which will be shown in our experiments.

\textbf{Momentum-based stochastic update.}
We propose a novel technique, momentum-based stochastic update, to optimize $\Delta z_i$.
We employ one-step optimization with using perturbations in previous iterations as momentum, i.e., $\Delta z_{i}$ will be initialized as $\Delta z_{i-1}$ in Equation~\eqref{eq:ftm_obj}.
Momentum can effectively improve the optimization effectiveness without increaseing additional optimization steps.
In addition, as we only require one-step optimization, the gradient for $\Delta z_i$ can be jointly computed with the gradient for $x^{\text{adv}}_i$, which can further speed up the optimization process.
To further improve the optimization efficiency, we use a stochastic update strategy, where only the randomly selected layers are updated in each attack iteration.
For each selected layer $k$, it has probability $p$ to be updated in the $i$-th attack iteration, otherwise, we set $\Delta z_{k,i} = \Delta z_{k,i-1}$.
We illustrate the detailed optimization process for FTM in the following.

The forward process for FTM in the $i$-th attack iteration is formulated as follows:
\begin{align}
    \bar{z}_{k,i}      &= {z}_{k,i} + \beta \| {z}_{k,i} \| \cdot \frac{ \Delta z_{k,i}}{\| \Delta z_{k,i} \| + \bar\epsilon}, \label{eq:ftm_bar} \\ 
    {z}^{\prime}_{k,i}      &= \begin{cases}(1 - \alpha_{k,i}) \odot \bar{z}_{k,i} + \alpha_{k,i} \odot {z}_{k,i}^c, & \tau_k < p, \\
    \bar{z}_{k,i}, & \text{else},\\
    \end{cases}
    \label{eq:ftm_perturb}
\end{align}
where $\beta \| {z}_{k,i} \|$ is the custom perturbation budget for our feature perturbation $\Delta z_{k,i}$.
$\beta$ is a scaling factor, and $\bar\epsilon$ is a small constant.
We mix the output feature ${z}_{k,i}$ and our attack-specific perturbation $\Delta z_{k,i}$ to obtain $\bar{z}_{k,i}$.
Then, we use clean features to augment $\bar {z}_{k,i}$ to get $z^{\prime}_{k,i}$, which is the final perturbed feature that will be fed into the $k$-th network layer.
These perturbed features are computed layer by layer during the forward process.
${z}_{k,i}^c$ is a random clean feature, and $\alpha_{k,i}$ is a vector representing channel-wise mixing ratio.
We randomly sample $\alpha_{k,i}$ from $\mathcal{U}(0, \alpha_{max})$.
In each iteration, we randomly sample $\tau_k$ from $\mathcal{U}(0, 1)$ to determine whether to update the $k$-th perturbation $\Delta z_{k,i}$ during the $i$-th attack iteration.
$p$ is the probability threshold.

The backward process for FTM in the $i$-th attack iteration is formulated as follows:
\begin{align}
    \Delta z_{i}^\prime &= \{(\Delta z_{k,i}, \tau_k) \mid k=1,\ldots,n \},\\
    g_{i+1}^\prime, g^{\Delta z^\prime}_{}                &= \nabla_{x^{\text{adv}}_i, \Delta z_i^\prime} \mathcal{L} \left( F (x^{\text{adv}}_i ; \Delta z_i^\prime), y_t \right), \label{eq:ftm_grad} \\
    \Delta z_{k,i+1}                                     &= \begin{cases}
        \Delta z_{k,i} + g^{\Delta z^\prime}_{k}, & \tau_k < p, \\
        \Delta z_{k,i}, & \text{otherwise},
    \end{cases} \label{eq:ftm_update}
\end{align}
where $\Delta z_{i}^\prime$ represents a set containing both perturbations $\Delta z_{k,i}$ and their corresponding random variables $\tau_k$ from the forward process, which determines whether each perturbation needs to be updated.
$g^{\Delta z^\prime}_{k}$ denotes the gradient with respect to the $k$-th perturbation $\Delta z_{k,i}$.
The threshold $p$ remains consistent with that used in the forward process.
We summarize our FTM attack in Algorithm~\ref{alg:ftm}.

\textbf{Efficiency analysis.}
FTM does not require complex operations or additional forward and backward passes. Therefore, FTM is theoretically efficient on both CNNs and ViTs. In contrast, SI, VT, Admix, and SU require additional forward and backward passes, while ODI requires a computationally expensive input transformation.

\textbf{Ensemble of surrogate variants (FTM-E).}
Observing that FTM can efficiently perturb the surrogate model, we propose to further enhance the attack effectiveness through an ensemble strategy.
We create multiple copies of the surrogate model and independently apply FTM to each copy, then ensemble them as a new surrogate model.
We empirically show that utilizing an ensemble of two perturbed surrogate models can substantially enhance the attack performance while maintaining reasonable computational costs.

\begin{algorithm}[t]
    \algnewcommand\algorithmicinput{\textbf{Input:}}
    \algnewcommand\Input{\item[\algorithmicinput]}
    \algnewcommand\algorithmicoutput{\textbf{Output:}}
    \algnewcommand\Output{\item[\algorithmicoutput]}
    \caption{Feature Tuning Mixup (FTM) Algorithm}
    \label{alg:ftm}
	\begin{algorithmic}[1]
		\Input Surrogate model $F$, 
               clean image $x$, 
               target label $y_t$, iteration number $T$, perturbation budget $\epsilon$,
               input transformation $\phi(\cdot)$ (RDI, etc.), 
               step size $\eta$,
               decay factor $\mu$
        \Input Mixing range $\alpha_{max}$, probability $p$,
               mixing ratio $\beta$
        \Output An adversarial example $x^{\text{adv}}$
        \State $x^{\text{adv}}_0=x, g_0=0$. 
        \State Initialize perturbation $\Delta z_0 = 0$
		\For{$i = 0 \rightarrow T-1$}:
            \State Extend $F(x^{\text{adv}}_i)$ to $F(\phi(x^{\text{adv}}_i))$
            \State Select layers to update and obtain $\Delta z_{i}^\prime$ from $\Delta z_{i}$
            \State Forward and backward pass to get $g_{i+1}^\prime$ and $g^{\Delta z^\prime}_{}$ using Equations~\eqref{eq:ftm_perturb} and \eqref{eq:ftm_grad}
            \State Obtain $\Delta z_{i+1}$ using Equation~\eqref{eq:ftm_update}
            \State Calculate $\tilde g_{i+1} = \mu \cdot g_{i} + g_{i+1}^\prime/\| g_{i+1}^\prime\|_1$
            \State Update example
		        $x^{\text{adv}}_{i+1} = x^{\text{adv}}_i - \eta \cdot \text{sign}(\tilde g_{i+1})$
            \State Clip $x^{\text{adv}}_{i+1}$ to be $\ell_{\infty}$-bounded by $\epsilon$
		\EndFor
        \State $x^{\text{adv}}=x^{\text{adv}}_T$
        \State \Return $x^{\text{adv}}$
	\end{algorithmic}
\end{algorithm}

\section{Experiments}
\label{sec:experiments}

\begin{table*}[!t]
    \centering
    \resizebox{0.96\textwidth}{!}{%
    \begin{tabular}{clc|ccccccccccc}
    \toprule[0.1em]
    Source & Attack & Time (s) & VGG-16 & RN-18 & RN-50 & DN-121 & Xcep & MB-v2 & EF-B0 & IR-v2 & Inc-v3 & Inc-v4 & \textbf{Avg.} \\ \midrule
    \multirow{10}{*}{RN-50} 
    & DI         & 1.37 & 63.2 & 56.4 & \textbf{98.9} & 74.7 & 5.5 & 27.8 & 29.0 & 5.6 & 9.0 & 9.4 & 37.9 \\
    & RDI        & 1.23 & 66.3 & 70.7 & 98.2 & 81.9 & 13.2 & 46.4 & 47.2 & 17.1 & 29.9 & 23.3 & 49.4 \\
    & RDI-SI     & 5.95 & 70.5 & 79.8 & \underline{98.8} & 88.9 & 29.5 & 56.2 & 66.2 & 37.9 & 56.4 & 43.6 & 62.8 \\
    & RDI-VT     & 7.16 & 68.8 & 78.7 & 98.2 & 82.5 & 27.9 & 54.5 & 56.1 & 32.8 & 45.8 & 37.9 & 58.3 \\
    & RDI-Admix  & 3.55 & 74.2 & 80.7 & 98.7 & 86.8 & 20.9 & 59.4 & 56.1 & 26.7 & 42.7 & 34.1 & 58.0 \\
    & RDI-SU     & 2.21 & 67.9 & 74.7 & 98.3 & 83.3 & 24.9 & 56.3 & 50.2 & 20.9 & 32.1 & 26.2 & 53.5 \\
    & ODI        & 4.38 & 78.3 & 77.1 & 97.6 & 87.0 & 43.8 & 67.3 & 70.0 & 49.5 & 65.9 & 55.4 & 69.2 \\
    & RDI-CFM    & 1.39 & 84.7 & \underline{88.4} & 98.4 & \underline{90.3} & 51.1 & 81.5 & 78.8 & 48.0 & 65.5 & 59.3 & 74.6 \\
    \cmidrule(l){2-14}
    & RDI-FTM    & 1.54 & \underline{86.3} & 87.5 & 97.9 & 89.7 & \underline{56.1} & \underline{83.3} & \underline{81.2} & \underline{54.7} & \underline{70.8} & \underline{66.6} & \underline{77.4} \\
    & RDI-FTM-E  & 2.92 & \textbf{88.1} & \textbf{88.6} & 98.3 & \textbf{91.4} & \textbf{59.4} & \textbf{85.4} & \textbf{84.3} & \textbf{56.9} & \textbf{73.4} & \textbf{69.3} & \textbf{79.5} \\
    \midrule
    \multirow{10}{*}{Inc-v3} 
   & DI        & 2.01 & 2.9 & 2.4 & 3.4 & 5.0 & 1.9 & 1.8 & 3.7 & 3.0 & \textbf{99.2} & 4.2 & 12.8 \\
   & RDI       & 1.76 & 3.5 & 3.8 & 4.0 & 7.0 & 3.1 & 3.0 & 5.9 & 6.3 & 98.7 & 7.1 & 14.2 \\
   & RDI-SI    & 8.11 & 4.0 & 5.2 & 5.7 & 11.0 & 6.3 & 4.6 & 8.2 & 11.6 & \underline{98.8} & 12.1 & 16.8 \\
   & RDI-VT    & 10.50 & 5.9 & 8.9 & 9.4 & 13.2 & 7.4 & 5.9 & 9.8 & 12.3 & 98.7 & 14.7 & 18.6 \\
   & RDI-Admix & 4.92 & 6.3 & 6.5 & 8.8 & 12.8 & 6.0 & 6.1 & 10.9 & 12.2 & 98.7 & 13.6 & 18.2 \\
   & RDI-SU    & 2.36 & 7.8 & 11.0 & 11.4 & 12.7 & 9.4 & 8.2 & 6.3 & 8.3 & 98.7 & 10.9 & 18.5 \\
   & ODI       & 6.42 & 14.3 & 14.9 & 16.7 & 32.3 & 20.3 & 13.7 & 25.3 & 26.4 & 95.9 & 31.6 & 29.1 \\
   & RDI-CFM   & 2.13 & 22.9 & 26.8 & 26.2 & 39.1 & 34.1 & 27.1 & 38.6 & 36.2 & 95.9 & 44.8 & 39.2 \\
    \cmidrule(l){2-14}
   & RDI-FTM & 2.35 & \underline{25.8} & \underline{30.9} & \underline{32.1} & \underline{41.7} & \underline{36.3} & \underline{30.7} & \underline{41.4} & \underline{42.9} & 94.1 & \underline{48.1} & \underline{42.4} \\
   & RDI-FTM-E & 4.37 & \textbf{34.2} & \textbf{39.3} & \textbf{40.6} & \textbf{54.4} & \textbf{47.0} & \textbf{37.1} & \textbf{51.6} & \textbf{52.4} & 96.6 & \textbf{59.2} & \textbf{51.2} \\
    \bottomrule[0.1em]
    \end{tabular}%
    }
    \caption{Targeted attack success rates (\%) against ten target models.
    All methods are combined with MI-TI.
    The best results are shown in bold, and the second best results are underlined.
    Time (s) denotes the average computation time required to attack a single image.
    }
    \label{tab:cnn}
    \vspace{-0.3cm}
\end{table*}

\subsection{Experimental Settings}
\label{sec:exp_settings}
\textbf{Datasets.} Following prior research~\cite{Byun_2022_CVPR,byun2023introducing,Wei_2023_CVPR,Zhao_2021_NeurIPS}, we conduct experiments on the ImageNet-compatible dataset from the NIPS 2017 adversarial attack challenge. This dataset comprises 1,000 images, each annotated with both true and target class labels for targeted attack evaluation.

\textbf{General settings.}
We adopt experimental configurations consistent with recent literature \cite{Zhao_2021_NeurIPS,Byun_2022_CVPR,byun2023introducing}. Specifically, we use an $\ell_\infty$-norm setting with $\epsilon=16/255$. We set decay $\mu=1.0$ and step size $\eta=2/255$. We perform 300 iterations ($T=300$) for iterative targeted attacks as per \cite{byun2023introducing}. For optimization of adversarial examples, all methods employ the simple logit loss function \cite{Zhao_2021_NeurIPS}.

\textbf{Models.}
We evaluate our method using 15 pre-trained DNNs, consisting of 10 CNN-based models and 5 transformer-based models.
For CNN-based architectures, we employ
VGG-16 \cite{SimonyanZ14a}, ResNet-18 (RN-18) \cite{He_2016_CVPR}, ResNet-50 (RN-50) \cite{He_2016_CVPR}, DenseNet-121 (DN-121) \cite{huang2017densely}, Xception (Xcep) \cite{chollet2017xception}, MobileNet-v2 (MB-v2) \cite{sandler2018mobilenetv2}, EfficientNet-B0 (EF-B0) \cite{tan2019efficientnet}, Inception ResNet-v2 (IR-v2) \cite{szegedy2017inception}, Inception-v3 (Inc-v3) \cite{SzegedyVISW16}, and Inception-v4 (Inc-v4) \cite{szegedy2017inception}.
For Transformer-based architectures, we utilize Vision Transformer (ViT) \cite{dosovitskiy2020vit}, LeViT \cite{graham2021levit}, ConViT \cite{d2021convit}, Twins \cite{chu2021twins}, and Pooling-based ViT (PiT) \cite{heo2021rethinking}.

\textbf{Baseline attacks.} 
We mainly compare our FTM against ten existing techniques: DI \cite{XieZZBWRY19}, RDI \cite{zou2020improving}, MI \cite{DongLPS0HL18}, TI \cite{DongPSZ19}, SI \cite{LinS00H20}, VT \cite{Wang021}, Admix \cite{Wang_2021_admix}, ODI \cite{Byun_2022_CVPR}, SU~\cite{Wei_2023_CVPR}, and CFM~\cite{byun2023introducing}. We primarily reference CFM, the current state-of-the-art method for targeted transfer-based attacks that requires no additional training dataset. MI and TI techniques apply across all attack methods, so we omit the `MI-TI' prefix in our references. RDI serves as a common baseline technique in most cases. For DI, RDI, TI, SI, Admix, and ODI, we follow the configurations specified in CFM. 
For Admix, we also follow CFM to set the mixing weight to $w=0.2$ and use $N=3$ mixed images (i.e., $m_2=3$ in \cite{Wang_2021_admix}).
Given Admix's computational intensity, we set the number of scale images in Admix to 1 (i.e., $m_1=1$ in \cite{Wang_2021_admix}) unless otherwise specified.
We report the results for ODI instead of RDI-ODI, as ODI performs similarly to RDI-ODI~\cite{Byun_2022_CVPR,byun2023introducing}.
For SU and CFM, we use the original settings from their papers.

\textbf{Settings for our FTM.}
We set the mixing ratio $\beta$ to 0.01.
For FTM-E, we set the number of perturbed models to 2 for all experiments due to the computational cost.
Following CFM, we avoid perturbing shallow layers because their low-level features vary signiﬁcantly with input transforms and may cause excessive disturbance during attacks, as explained in CFM’s paper.
Specifically, we perturb layers where the feature map size is smaller than or equal to $\frac{1}{16}$ of the image size.
For a pure ViT, we select the outputs of all fully-connected layers.
We sample $\bm{\alpha}_{k,i}$ from $\mathcal{U}(0, 0.75)$, i.e., $\alpha_{max}=0.75$.
We set the probability $p$ to $0.1$.

\subsection{Experimental Results}

\begin{table}[!t]
    \centering
    \resizebox{0.48\textwidth}{!}{%
    \begin{tabular}{clc|cccccc}
     \toprule[0.1em]
    Source & Attack & Time (s) & ViT & LeViT & ConViT & Twins & PiT & \textbf{Avg.} \\
    \midrule
    \multirow{10}{*}{RN-50}
    & DI        & 1.37 &  0.2 & 3.5 & 0.4 & 1.5 & 1.7 & 1.5       \\
    & RDI       & 1.23 &  0.7 & 13.2 & 1.7 & 6.1 & 7.0 & 5.7      \\
    & RDI-SI    & 5.95 &  2.9 & 29.4 & 6.3 & 15.5 & 17.9 & 14.4   \\
    & RDI-VT    & 7.16 &  2.9 & 28.1 & 5.2 & 15.0 & 14.0 & 13.0   \\
    & RDI-Admix & 3.55 &  1.3 & 22.5 & 2.5 & 8.5 & 8.4 & 8.6      \\
    & RDI-SU    & 2.21 &  0.8 & 16.9 & 2.3 & 6.9 & 7.8 & 6.9      \\
    & ODI       & 4.38 &  5.1 & 37.0 & 10.7 & 20.1 & 29.1 & 20.4  \\
    & RDI-CFM   & 1.39 &  4.3 & 46.1 & 8.9 & 25.2 & 24.7 & 21.8   \\
    \cmidrule(l){2-9}
    & RDI-FTM   & 1.54 & \underline{5.9} & \underline{52.9} & \underline{10.8} & \underline{32.4} & \underline{31.5} & \underline{26.7} \\
    & RDI-FTM-E & 2.92 & \textbf{6.8} & \textbf{58.6} & \textbf{13.6} & \textbf{35.2} & \textbf{34.9} & \textbf{29.8} \\
    \midrule
    \multirow{10}{*}{Inc-v3}
    & DI        & 2.01 & 0.1 & 0.3 & 0.0 & 0.0 & 0.1 & 0.1     \\
    & RDI       & 1.76 & 0.2 & 1.8 & 0.2 & 0.4 & 0.7 & 0.7     \\
    & RDI-SI    & 8.11 & 0.3 & 4.1 & 0.9 & 0.7 & 3.2 & 1.8     \\
    & RDI-VT    & 10.5 & 0.4 & 5.2 & 0.8 & 1.6 & 1.8 & 2.0     \\
    & RDI-Admix & 4.92 & 0.1 & 4.1 & 0.6 & 1.4 & 1.4 & 1.5     \\
    & RDI-SU    & 2.36 & 0.2 & 2.0 & 0.3 & 1.3 & 1.0 & 1.0     \\
    & ODI       & 6.42 & 0.8 & 12.4 & 1.7 & 3.5 & 6.7 & 5.0    \\
    & RDI-CFM   & 2.13 & 2.1 & 21.9 & 3.2 & 6.1 & 11.6 & 8.9   \\
    \cmidrule(l){2-9}
    & RDI-FTM   & 2.35 & \underline{2.4} & \underline{25.0} & \underline{4.5} & \underline{10.0} & \underline{15.3} & \underline{11.5} \\
    & RDI-FTM-E & 4.37 & \textbf{3.8} & \textbf{32.7} & \textbf{6.8} & \textbf{12.7} & \textbf{20.4} & \textbf{15.3} \\
    \bottomrule[0.1em]
    \end{tabular}
    
    }
    \caption{Targeted attack success rates (\%) against five transformer-based DNNs. 
    All methods are combined with MI-TI.
    The best results are shown in bold, and the second best are underlined.
    Time (s) denotes the average time required to attack a single image.
    }  
    \vspace{-0.1cm}
    \label{tab:vit}
\end{table}

\textbf{Transfer success rates.}
Table~\ref{tab:cnn} shows the transfer success rates of various attack methods on ten black-box models.
When using Inc-v3 as the source model, RDI-FTM-E achieves an average success rate of 51.2\%, surpassing the previous best method CFM by 12.0\%. 
The improvement is particularly notable for challenging target models like IR-v2 and Inc-v4, where RDI-FTM-E increases the success rates by 16.2\% and 14.4\%, respectively.
When using RN-50 as the source model, both RDI-FTM and RDI-FTM-E achieve higher success rates than CFM.
The results show that the effective feature-level augmentation employed by FTM helps generate transferable adversarial examples.

Table~\ref{tab:vit} shows the transfer success rates of various attack methods on five transformer-based models.
Our methods also achieve significant improvements over existing methods.
For instance,
when using RN-50 as the source model, RDI-FTM and RDI-FTM-E improve the targeted attack success rates of CFM by 4.9\% and 8.0\%, respectively.
The results suggest that our method can effectively improve the transferability of adversarial examples across different architectures.

\begin{figure}[t]
    \centering
    \setlength{\belowcaptionskip}{-0.2cm}
     \includegraphics[width=\linewidth]{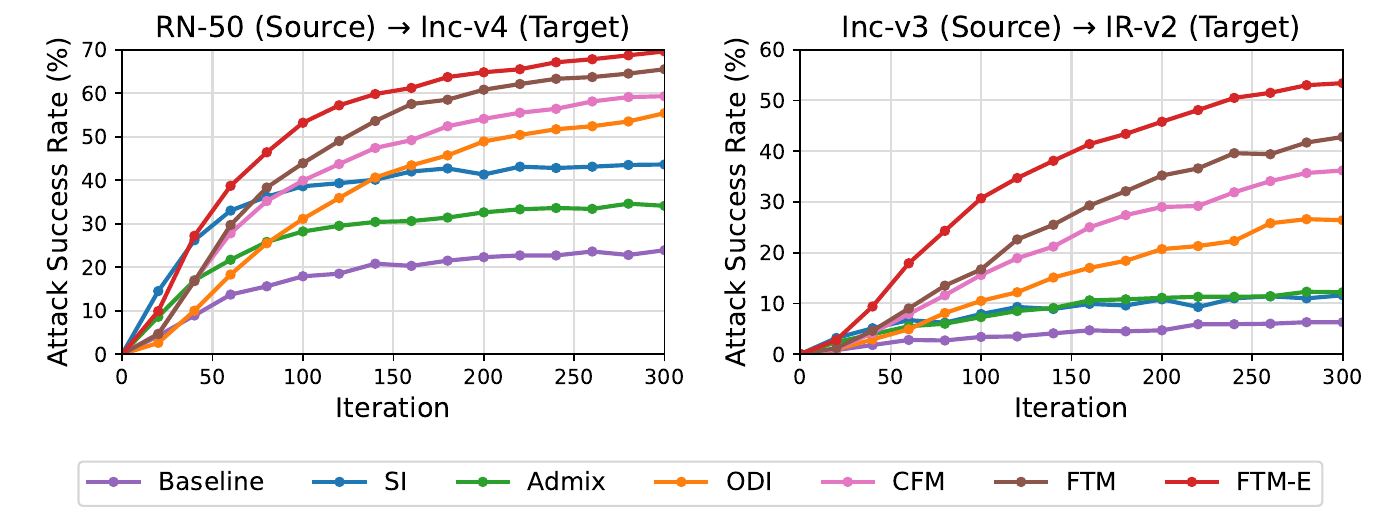}
     \caption{
      Targeted attack success rates (\%) based on the number of iterations.
      The source model is RN-50 and Inc-v3, respectively.
     }
     \label{fig:iter}
     \vspace{-0.1cm}
  \end{figure}

Figure~\ref{fig:iter} shows the targeted attack success rates based on the number of iterations.
We evaluate our methods under different numbers of iterations using two experimental settings: RN-50 $\rightarrow$ Inc-v4 and Inc-v3 $\rightarrow$ IR-v2. 
The results show that FTM and FTM-E achieve stable performance across different iteration counts.
For the RN-50 $\rightarrow$ Inc-v4 setting, SI exhibits superior performance at the beginning of the optimization process.
However, as the number of iterations increases, FTM surpasses existing approaches.
In the Inc-v3 $\rightarrow$ IR-v2 setting, FTM-E significantly outperforms other methods across the entire range of iterations.

\begin{figure}[t]
  \centering
  \includegraphics[width=\linewidth]{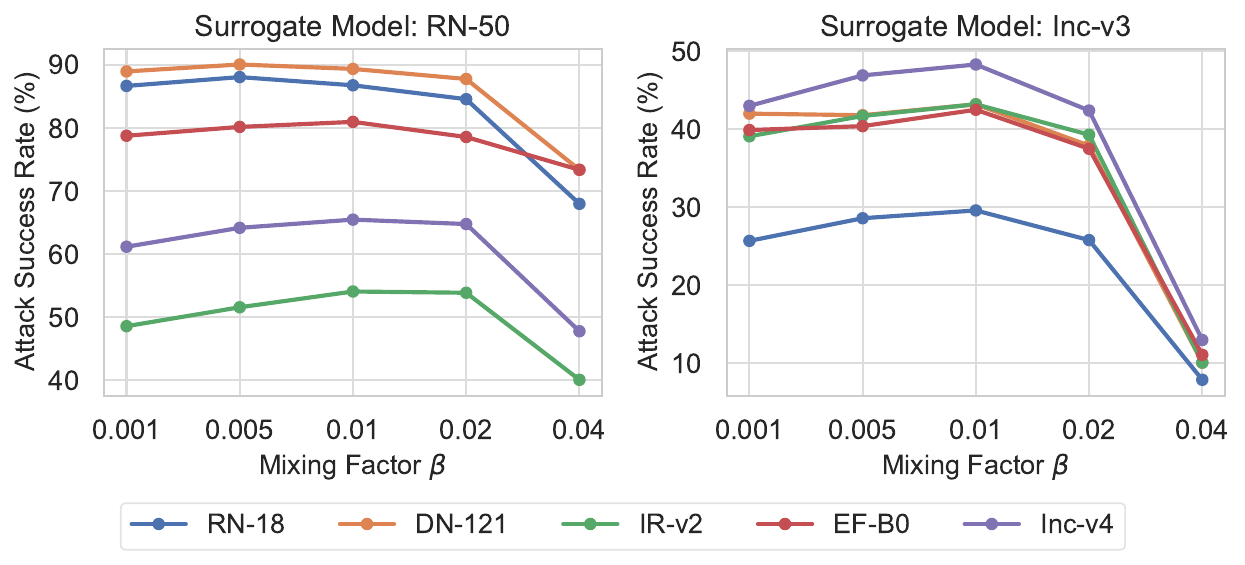}
  \setlength{\abovecaptionskip}{-0.2cm}
  \setlength{\belowcaptionskip}{-0.2cm}
   \caption{
  Attack success rates with different mixing ratio $\beta$.
  Each color represents the attack effectiveness on a specific target model.
   }
   \label{fig:beta}
\end{figure}


\begin{table*}[!t]
    \centering
    \resizebox{0.94\textwidth}{!}{%

    \begin{tabular}{lc|cccccccccccccc}
        \toprule[0.1em]
         \multirow{2}{*}{Attack}  & \multicolumn{14}{c}{RN-50 + Inc-v3 $\Rightarrow$} &  \\\cmidrule(l){2-15}
                    & Time (s) & VGG-16 & RN-50* & Inc-v3* & DN-121 & IR-v2 & Inc-v4 & Xcep & ViT & LeViT & ConViT & Twins & PiT & \textbf{Avg.} \\
        \midrule
        RDI & 3.26 &66.9 & \textbf{98.4} & \textbf{98.7} & 83.1 & 35.6 & 42.5 & 25.4 & 0.8 & 22.4 & 3.2 & 9.8 & 10.8 & 41.5 \\
        ODI & 7.17 & 70.9 & 94.0 & 96.1 & 83.4 & 65.4 & 69.9 & 57.6 & 9.1 & 52.9 & 18.6 & 28.6 & 42.8 & 57.4 \\
        RDI-CFM & 4.39 & 86.1 & \underline{98.3} & 97.0 & 92.1 & 71.7 & 79.9 & 72.0 & 10.3 & 66.5 & 20.3 & 41.9 & 45.8 & 65.2 \\
        RDI-ODI-CFM & 7.73 & 63.9 & 88.6 & 81.3 & 74.1 & 54.3 & 56.0 & 54.3 & 12.6 & 49.9 & 20.9 & 29.6 & 41.0 & 52.2 \\
        RDI-Admix-CFM & 13.38 & 86.2 & 98.1 & \underline{97.7} & 92.7 & 75.1 & 81.6 & 76.1 & 14.0 & 73.1 & 24.8 & 50.7 & 53.9 & 68.7 \\
        RDI-SI-CFM & 22.13 & 87.5 & 98.1 & 97.3 & 92.7 & 78.0 & 81.6 & 78.0 & 20.8 & 76.3 & 32.9 & 55.0 & 59.3 & 71.5 \\
        \rowcolor{Gray} RDI-FTM & 4.47 & 87.5 & 97.8 & 96.7 & 92.4 & 75.4 & 81.7 & 74.0 & 14.5 & 71.0 & 25.2 & 49.3 & 54.4 & 68.3 \\
        \rowcolor{Gray} RDI-FTM-E & 8.81 & \textbf{88.2} & 98.1 & 97.1 & \underline{93.4} & 76.7 & 83.4 & 77.1 & {18.4} & {74.9} & {30.1} & {53.4} & {58.0} & 70.7 \\
        \rowcolor{Gray} RDI-Admix-FTM & 14.10 & 86.0 & 96.7 & 96.5 & 92.2 & 78.3 & \textbf{83.8} & 77.3 & 20.1 & 75.8 & 29.9 & 55.7 & 58.5 & 70.9 \\
        \rowcolor{Gray} RDI-SI-FTM & 22.66 & 87.5 & 96.8 & 96.2 & 93.3 & \underline{79.0} & \underline{83.6} & \textbf{79.1} & \underline{25.7} & \underline{77.7} & \underline{37.5} & \underline{59.9} & \underline{64.3} & \underline{73.4} \\
        \rowcolor{Gray} RDI-SI-FTM-E & 43.43 & \underline{87.8} & 97.3 & 96.7 & \textbf{93.9} & \textbf{79.1} & 83.5 & \underline{78.2} & \textbf{27.3} & \textbf{79.2} & \textbf{42.9} & \textbf{62.9} & \textbf{65.6} & \textbf{74.5} \\

    \bottomrule[0.1em]
    \end{tabular}
    }
    \caption{
        Targeted attack success rates (\%) of combination with different methods.
        The surrogate model is the ensemble of RN-50 and Inc-v3.
        All methods are combined with MI-TI.
        The best results are shown in bold, and the second best are underlined. 
        Time (s) denotes the average time required to attack a single image.
    }  
    \vspace{-0.4cm}
    \label{tab:combine}
\end{table*}

\textbf{Computational overhead.}
We analyze the computational efficiency of different methods and report their costs in Table~\ref{tab:vit}.
We use an NVIDIA 3090 GPU to compute the time per image.
The results demonstrate that our approach introduces reasonable computational overhead compared to CFM.
Specifically, with RN-50 as the source model, RDI-FTM requires an average of 1.54 seconds per image, representing only a 10.8\% increase over CFM's 1.39 seconds.
In contrast, alternative methods like SI, VT, and ODI impose substantially higher computational burdens.
For instance, VT requires 7.16 seconds per image, and SI requires 5.95 seconds per image.
We also include relevant analysis in Figure~\ref{fig:starter}.
The result shows that FTM achieves strong attack performance while maintaining low computational cost.

\textbf{Combination with existing techniques.}
Table~\ref{tab:combine} presents transfer success rates of our method combined with ensemble-based method~\cite{LiuCLS17}, SI, and Admix.
Using the ensemble of RN-50 and Inc-v3 as the surrogate model, our RDI-SI-FTM-E achieves 74.5\% average success rate, surpassing existing methods.
FTM notably enhances performance on transformer-based models.
For instance, RDI-SI-FTM achieves 64.3\% on PiT, which is 5.0\% higher than RDI-SI-CFM (59.3\%) with comparable computational cost.
These results demonstrate that FTM is a generic technique that combines effectively with existing attack methods.

\subsection{Ablation Study}
\label{sec:ablation}

For all ablation experiments, we use RN-50 to generate adversarial examples and evaluate the average success rates on all models from Section~\ref{sec:exp_settings}, unless stated otherwise.

\begin{figure}[t]
  \centering
  \includegraphics[width=\linewidth]{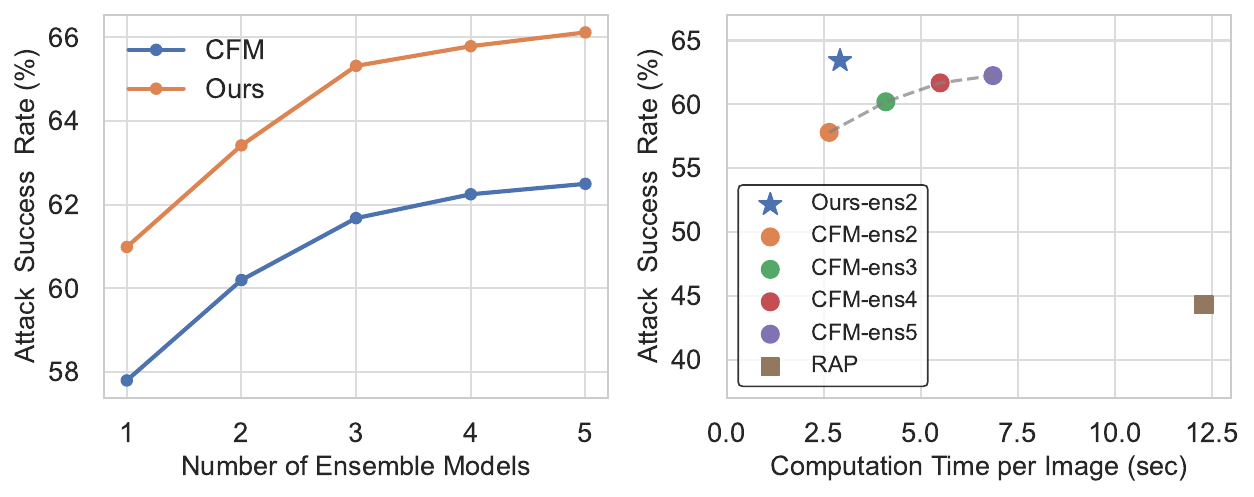}
  \setlength{\abovecaptionskip}{-0.3cm}
  \setlength{\belowcaptionskip}{-0.2cm}
   \caption{
  Attack performance and computational costs with different numbers of ensemble models.
  Left: Our FTM vs. CFM using ensemble strategy.
  Right: Computational costs of our FTM and RAP. -ens2 means using 2 perturbed models, and so on.
   }
   \label{fig:ensemble_ablation}
\end{figure}

\begin{figure}[t]
  \centering
  \includegraphics[width=\linewidth]{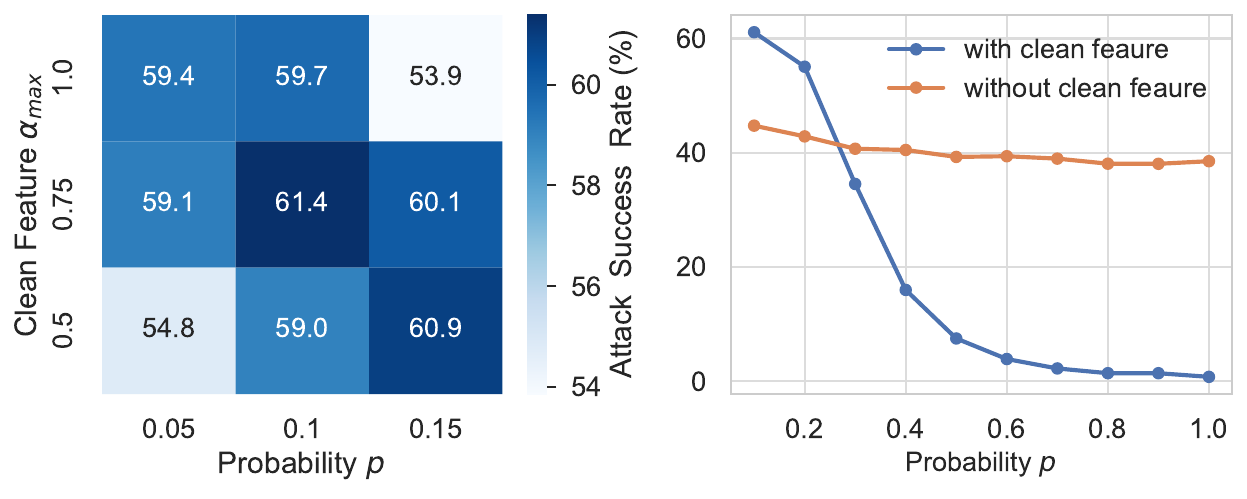}
  \setlength{\abovecaptionskip}{-0.3cm}
  \setlength{\belowcaptionskip}{-0.2cm}
   \caption{
  Left: Attack success rates vs. $p$ and $\alpha_{max}$. 
  Right: Attack success rates vs. $p$ with/without clean features;
  The blue line is the success rates when the clean features from CFM are integrated with our FTM, while the orange line represents only using our learnable perturbations.
   }
   \label{fig:alpha_p_ablation}
   \vspace{-0.1cm}
\end{figure}

\textbf{Stability with respect to $\beta$.}
Figure~\ref{fig:beta} shows the transfer success rates of FTM across different mixing ratios $\beta$. The success rates on five target models are plotted in distinct colors. 
When $\beta$ exceeds 0.04, the success rates decrease substantially across all target models. However, for $\beta$ values below 0.02, we observe stable attack performance across all models. The results suggest that FTM maintains robust performance within a reasonable range of $\beta$ values, showing its stability with respect to $\beta$.

\textbf{Compare FTM and CFM using ensemble strategy.}
Figure~\ref{fig:ensemble_ablation} shows the attack performance of FTM and CFM using different numbers of perturbed models.
On the left, our FTM consistently outperforms CFM across all ensemble sizes.
Notably, FTM with only 2 perturbed models achieves better attack performance than an ensemble of 5 CFM-perturbed models, underscoring FTM's superior effectiveness.
It is also worth mentioning that single FTM and CFM have comparable time complexity (Table~\ref{tab:vit}).
On the right, we compare our FTM with RAP~\cite{qin2022boosting}, and the results show that FTM significantly outperforms RAP in both effectiveness and computational efficiency.

\textbf{Analysis of probability $p$ and ratio $\alpha_{max}$.}
Figure~\ref{fig:alpha_p_ablation} presents an analysis of FTM's attack success rates under varying values of $p$ and $\alpha_{max}$.
The left plot shows that FTM achieves optimal performance with $p=0.1$ and $\alpha_{max}=0.75$.
FTM maintains competitive performance under various parameter configurations, such as $p,\alpha_{max}=0.15,0.5$ or $0.05,1.0$, indicating its parameter stability.
The right plot illustrates the impact of clean features on FTM. With clean features enabled, FTM’s performance declines significantly as $p$ increases from 0.1 to 1.0, while FTM without clean features shows stable performance across varying $p$ values. This suggests that although clean features are crucial to FTM’s performance, they also add instability to the attack process. 
Additionally, setting $p=1.0$ (where all layers are updated) causes excessive perturbations, making the attack overly aggressive and reducing its effectiveness.

The drop of the blue line in Figure~\ref{fig:alpha_p_ablation} occurs because the clean accuracy of the perturbed surrogate model becomes extremely low (3.8\%) at $p=1.0$ when using clean feature mixup. Without clean feature mixup (orange line), the accuracy remains high at 65.3\%. The low accuracy makes it impossible to generate effective adversarial examples, causing the attack success rate of the blue line to drop to near 0.

\section{Conclusion}
In this work, we study the challenging problem of enhancing transferability in targeted adversarial attacks.
Our investigation reveals that feature-level augmentation strategies show promising potential for boosting transferability effectively and efficiently. 
We introduce Feature Tuning Mixup (FTM), a novel method that incorporates both attack-specific feature perturbations and clean features in the feature space. To optimize the learnable perturbations efficiently, we develop a momentum-based stochastic update strategy.
Comprehensive experiments on the ImageNet-compatible dataset demonstrate that FTM achieves substantial improvement over current state-of-the-art methods while maintaining low computational cost.

\vspace{0.15cm}
\noindent
\textbf{Acknowledgments.}
This work was supported in part by HK RGC GRF under Grant PolyU 15201323.

\clearpage

{
    \small
    \bibliographystyle{ieeenat_fullname}
    \bibliography{sample-base}
}


\clearpage
\clearpage

\twocolumn[{
\begin{center}
    {\Large \textbf{Supplementary Material for}} \\[0.6em] 
    {\Large \textbf{Improving Transferable Targeted Attacks with Feature Tuning Mixup}}
    \vspace{7ex}
\end{center}
}]

\renewcommand{\thesection}{\Alph{section}}
\setcounter{section}{0} 

In the supplementary material, we provide additional experimental results on parameter $p$ in Chapter~\ref{sec:supp-para-p}. Then, in Chapters~\ref{sec:LLMs},~\ref{sec:dn121-levit}, and~\ref{sec:efficiency-vit}, we present additional results demonstrating the effectiveness and efficiency of FTM across different surrogate and targeted models. Finally, Chapter~\ref{sec:Visualization} includes visualizations of adversarial examples generated by our FTM.

\section{Additional Experiments}
\label{sec:additional_results}

\subsection{Supplementary results on the analysis of $p$}
\label{sec:supp-para-p}

We have shown in Section~\ref{sec:ablation} (Ablation Study) that the attack success rate when using clean feature mixup drops to near $0$ in Figure~\ref{fig:alpha_p_ablation} due to the low clean accuracy of the perturbed surrogate model.
Table~\ref{tab:p} shows the clean accuracy of the perturbed surrogate model with varying values of $p$.
The results demonstrate that increasing $p$ leads to extremely low clean accuracy when using clean feature mixup in FTM,
which explains the rapid decline of the blue line in Figure~\ref{fig:alpha_p_ablation}.
In contrast, without clean feature mixup, the clean accuracy remains stable, maintaining 65.3\% even at $p = 1.0$.
Consequently, the corresponding orange line in Figure~\ref{fig:alpha_p_ablation} exhibits relatively stable behavior. 
Notably, as shown in the left plot of Figure~\ref{fig:alpha_p_ablation} our FTM performs stably around the optimal hyperparameter $p$ and $\alpha_
{max}$.

\begin{table}[!h]
    \centering
      \centering
      \resizebox{\linewidth}{!}{%
      \begin{tabular}{lcccccccccc}
      \toprule[0.15em]
       \multirow{2}{*}{Ablation} & \multicolumn{10}{c}{Parameter $p$} \\
       \cmidrule(l){2-11}
       & 0.1 & 0.2 & 0.3 & 0.4 & 0.5 & 0.6 & 0.7 & 0.8 & 0.9 & 1.0\\
      \midrule
      w/ clean feature mixup  & 51.9 & 28.3 & 14.0 & 10.8 & 8.2 & 6.0 & 5.4 & 4.9 & 4.3 & 3.8\\
      w/o clean feature mixup & 79.8 & 73.3 & 69.7 & 68.6 & 67.0 & 66.4 & 66.0 & 65.8 & 65.4 & 65.3\\
      \bottomrule[0.15em]
      \end{tabular}%
      }
      \caption{
        Accuracy (\%) of clean images on the perturbed surrogate model with varying parameter $p$.
      }
      \label{tab:p}
\end{table}

\subsection{Attacking black-box multimodal LLMs}
\label{sec:LLMs}

FTM can be used to attack multimodal LLMs.
We use four commercial LLMs for evaluation, including Qwen2-VL, Llama-3.2, Claude-3.5, and GPT-4o. 
We randomly select 100 images from the ImageNet-compatible dataset and generate targeted adversarial examples on ViT using RDI-FTM-E.
For each generated targeted adversarial example, we use the prompt “Is this image a photo of \{target label\}? Yes or No?” to obtain the predictions of the LLMs. Table~\ref{tab:MLLMs} shows that our method achieves an average attack success rate of 40.5\%.

\begin{table}[!h]
    \centering
    \resizebox{\linewidth}{!}{%
    \begin{tabular}{lcccc|cccc}
    \toprule[0.15em]
    Response & Qwen2-VL & Llama-3.2 & Claude-3.5 & GPT-4o & Avg \\
  
    \midrule
    Total            & 100 & 100 & 100 & 100 & 100 \% \\
    Refuse to Answer & 0   & 10  & 0 & 0 & 2.50 \% \\
    Uncertain        & 1   & 5   & 1 & 0 & 1.75 \% \\
    Attack Failed    & 52  & 42  & 54 & 73 & 55.25 \% \\
    \rowcolor{Gray}
    Attack Succeeded & 47  & 43  & 45 & 27 & 40.50 \% \\

    \bottomrule[0.15em]
    \end{tabular}%
    }
    \caption{
    Evaluation on multimodal LLMs.
    The targeted adversarial examples are generated on ViT using RDI-FTM-E.
    }
    \label{tab:MLLMs}
  \end{table}

\subsection{Evaluation with different surrogate models}
\label{sec:dn121-levit}

We report the targeted attack success rates when using DN-121 or LeViT as the surrogate model in Table~\ref{tab:dn121_levit}.
The results demonstrate that, across all black-box attack scenarios, our methods consistently outperform existing approaches, regardless of whether DN-121 or LeViT is used as the surrogate model.

We report the targeted attack success rates using ensemble-based surrogate models in Table~\ref{tab:ensemble}.
We use two settings for the ensemble-based surrogate models: RN-50 + Inc-v3 and RN-50 + LeViT.
The results show that our methods can be combined with current ensemble-based methods to further improve the attack success rates.

\subsection{Efficiency of FTM on attacking ViT}
\label{sec:efficiency-vit}

We show that FTM remains effective and efficient when using ViT as the surrogate model. Unlike other models in Section~\ref{sec:exp_settings}, ViT’s feature map size does not decrease with depth. As analyzed in Section~\ref{sec:ftm-alg}, FTM is theoretically efficient and adaptable to various architectures, including ViT.
Tables~\ref{tab:vit-cnn} and~\ref{tab:vit-vit} confirm that FTM maintains low computational complexity when using ViT as the surrogate model. Our RDI-FTM-E-SI$_{m_1=2}$ generates adversarial examples in just 3.82 seconds on average while significantly outperforming existing attacks. In RDI-FTM-E-SI$_{m_1=2}$, the two copies of the surrogate model in FTM-E use scaled inputs (1 and 0.5, respectively), preserving efficiency.
The parameter $m_1$ in RDI-FTM-E-SI$_{m_1=2}$ is set to 2 to correspond with the two copies of the surrogate model used in FTM-E.
Overall, our results validate FTM’s effectiveness and efficiency across different architectures.

\subsection{Visualization of targetd adversarial examples}
\label{sec:Visualization}

\begin{figure*}[h]
    \centering
     \includegraphics[width=0.95\linewidth]{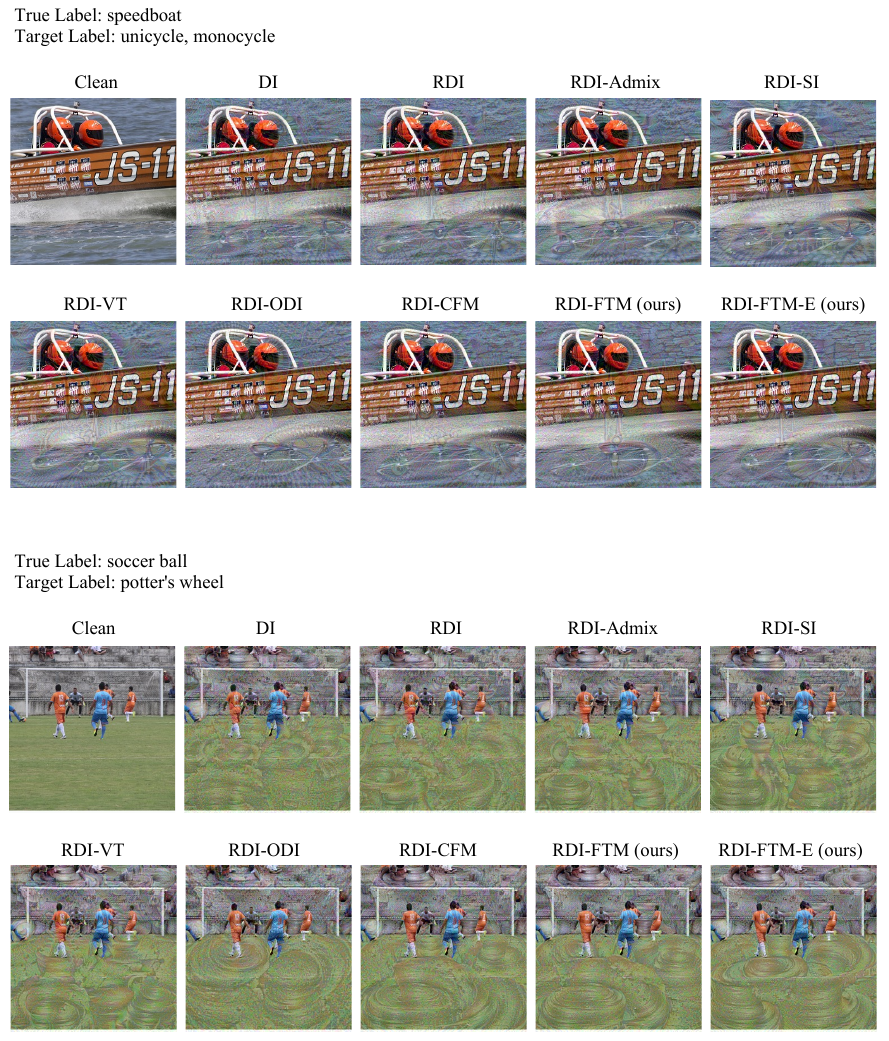}
  
     \caption{
        Visualization of targeted adversarial examples generated by different attack methods. The surrogate model used for attack generation is an ensemble of RN-50 and Inc-v3.
     }
     \label{fig:rn50_incv3_vis}
 \end{figure*}

\begin{figure*}[h]
    \centering
     \includegraphics[width=0.95\linewidth]{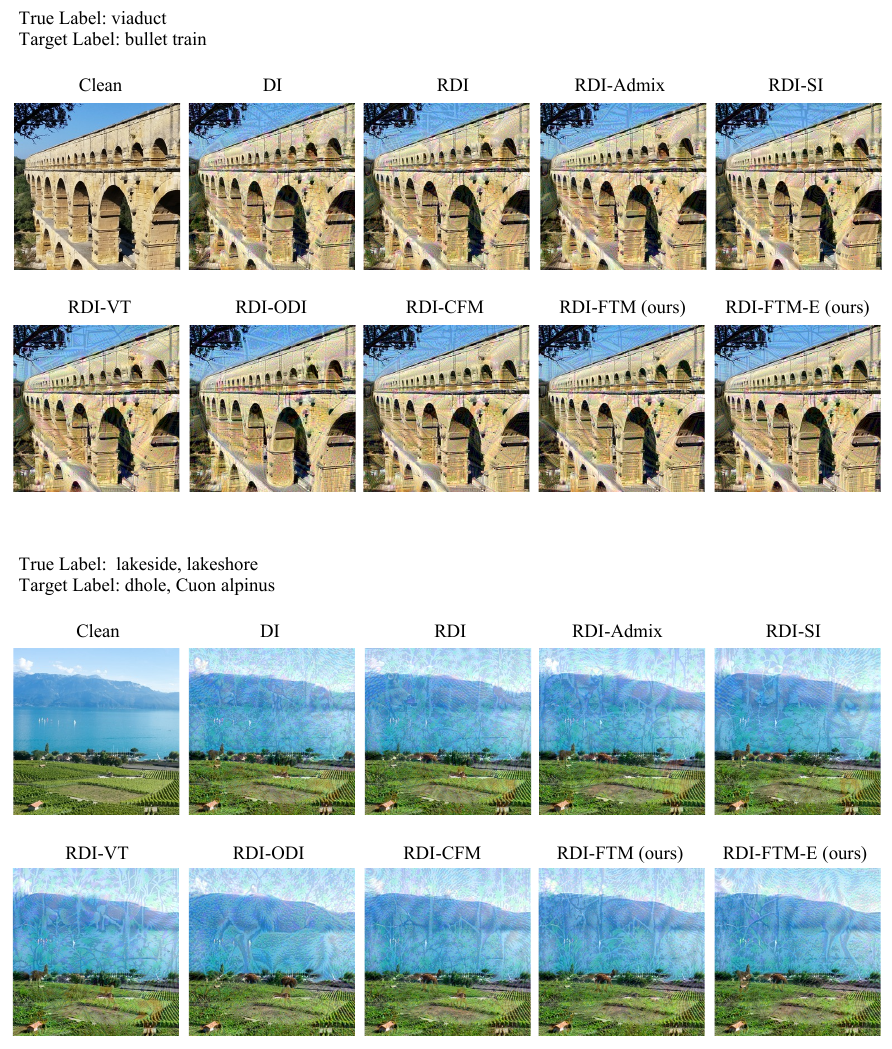}
  
     \caption{
        Visualization of targeted adversarial examples generated by different attack methods. The surrogate model used for attack generation is an ensemble of RN-50 and LeViT.
     }
     \label{fig:rn50_levit_vis}
 \end{figure*}

Figure~\ref{fig:rn50_incv3_vis} and Figure~\ref{fig:rn50_levit_vis} present visualizations of adversarial examples crafted by different targeted transfer-based attack methods.

\begin{table*}[!b]
    \centering
    \resizebox{0.9\textwidth}{!}{%
    \begin{tabular}{lccccccccccccc}
    \toprule[0.15em]
     \multirow{2}{*}{Attack}  & \multicolumn{12}{c}{DN-121 $\Rightarrow$} &  \\\cmidrule(l){2-13}
                & VGG-16 & RN-50 & Inc-v3 & DN-121* & IR-v2 & Inc-v4 & Xcep & ViT & LeViT & ConViT & Twins & PiT \\
    \midrule

    DI            & 37.1 & 44.4 & 7.1 & \textbf{98.7} & 4.3 & 8.3 & 5.2 & 0.2 & 3.0 & 0.4 & 1.0 & 1.1 \\           
    RDI           & 42.1 & 55.7 & 20.8 & 98.5 & 12.8 & 18.8 & 10.1 & 0.8 & 8.5 & 1.3 & 3.7 & 4.5 \\       
    RDI-Admix     & 53.2 & 67.6 & 31.1 & 98.3 & 20.1 & 26.5 & 17.8 & 1.0 & 14.7 & 1.7 & 6.8 & 7.4 \\      
    RDI-Admix$_5$ & 49.6 & 65.3 & 41.0 & \underline{98.6} & 28.9 & 34.3 & 21.6 & 2.4 & 21.8 & 3.4 & 10.5 & 14.2 \\    
    RDI-SI        & 45.4 & 60.1 & 34.3 & \underline{98.6} & 22.0 & 25.8 & 16.1 & 2.0 & 16.1 & 2.4 & 8.2 & 11.7 \\     
    RDI-VT        & 47.7 & 62.1 & 31.5 & \underline{98.6} & 25.4 & 27.2 & 20.3 & 2.2 & 19.2 & 3.5 & 8.3 & 11.7 \\     
    RDI-ODI       & 64.2 & 71.7 & 52.8 & 98.0 & 39.8 & 45.9 & 31.4 & 3.3 & 26.9 & 7.4 & 14.7 & 21.9 \\    
    RDI-CFM       & 76.2 & 83.9 & 56.1 & 97.8 & 43.6 & 53.8 & 41.1 & 3.6 & 32.8 & 6.4 & 17.3 & 21.1 \\    
    \cmidrule(l){1-13}

    \rowcolor{Gray} RDI-FTM     & \underline{77.3} & \underline{85.6} & \underline{62.8} & 98.2 & \underline{46.9} & \underline{58.4} & \underline{46.0} & \underline{3.7} & \underline{40.0} & \underline{8.2} & \underline{21.6} & \underline{26.5} \\
    \rowcolor{Gray} RDI-FTM-E   & \textbf{79.5} & \textbf{87.6} & \textbf{64.6} & 98.0 & \textbf{48.7} & \textbf{60.1} & \textbf{47.6} & \textbf{4.2} & \textbf{43.4} & \textbf{8.9} & \textbf{24.4} & \textbf{26.7} \\

    \toprule[0.15em]
    \addlinespace[0.4cm]
    \toprule[0.15em]
     \multirow{2}{*}{Attack}  & \multicolumn{12}{c}{LeViT $\Rightarrow$} &  \\\cmidrule(l){2-13}
                & VGG-16 & RN-50 & Inc-v3 & DN-121 & IR-v2 & Inc-v4 & Xcep & ViT & LeViT* & ConViT & Twins & PiT \\
    \midrule
    DI            & 1.6 & 2.5 & 4.2 & 2.7 & 1.8 & 2.6 & 2.3 & 0.6 & \textbf{100} & 4.2 & 9.3 & 10.3 \\           
    RDI           & 3.0 & 3.5 & 5.4 & 4.1 & 3.6 & 4.6 & 2.3 & 1.1 & \textbf{100} & 7.9 & 13.8 & 22.1 \\            
    RDI-Admix     & 6.5 & 8.3 & 9.9 & 8.8 & 5.8 & 7.2 & 5.5 & 3.3 & \textbf{100} & 10.7 & 23.3 & 30.9 \\         
    RDI-Admix$_5$ & 5.3 & 8.1 & 13.9 & 12.4 & 9.9 & 8.4 & 7.2 & 8.0 & 99.9 & 20.6 & 30.0 & 47.2\\                                                                            
    RDI-SI        & 3.4 & 6.3 & 10.6 & 10.0 & 6.4 & 5.4 & 5.1 & 4.3 & \textbf{100} & 18.1 & 24.1 & 38.4 \\         
    RDI-VT        & 5.3 & 7.2 & 12.0 & 10.1 & 8.3 & 8.8 & 8.9 & 6.3 & 99.9 & 18.7 & 27.2 & 40.5 \\        
    RDI-ODI       & 21.0 & 25.0 & 40.9 & 38.3 & 25.7 & 31.2 & 26.3 & 15.3 & 98.7 & 34.1 & 43.8 & 66.1 \\    
    RDI-CFM       & 27.3 & 30.3 & 39.8 & 39.0 & 23.6 & 30.1 & 27.2 & 18.4 & \textbf{100} & 45.5 & 63.8 & 75.7 \\   
                    
    \cmidrule(l){1-13}
    \rowcolor{Gray} RDI-FTM     & \underline{41.3} & \underline{41.9} & \underline{56.9} & \underline{54.2} & \underline{39.8} & \underline{46.5} & \underline{42.2} & \underline{31.1} & 99.9 & \underline{63.2} & \underline{77.5} & \underline{86.7} \\
    \rowcolor{Gray} RDI-FTM-E   & \textbf{50.1} & \textbf{52.4} & \textbf{62.8} & \textbf{63.1} & \textbf{48.4} & \textbf{53.2} & \textbf{49.2} & \textbf{40.3} & 99.9 & \textbf{72.2} & \textbf{86.2} & \textbf{93.0} \\ 

    \bottomrule[0.15em]
    \end{tabular}%
    }
    \caption{
        Targeted attack success rates (\%) using DN-121 or LeViT as surrogate model.
        All methods are combined with MI-TI.
        The best results are shown in bold, and the second best results are underlined.
        The surrogate models are marked with * in the first column.
    }
    \label{tab:dn121_levit}
\end{table*}

\begin{table*}[b]
    \centering
    \resizebox{0.9\textwidth}{!}{%
    \begin{tabular}{lccccccccccccc}
    \toprule[0.15em]
     \multirow{2}{*}{Attack}  & \multicolumn{12}{c}{RN-50 + Inc-v3 $\Rightarrow$} &  \\\cmidrule(l){2-13}
                & VGG-16 & RN-50* & Inc-v3* & DN-121 & IR-v2 & Inc-v4 & Xcep & ViT & LeViT & ConViT & Twins & PiT \\
    \midrule
    DI & 63.5 & \textbf{99.0} & \textbf{99.2} & 77.0 & 16.6 & 24.0 & 12.5 & 0.2 & 7.3 & 0.7 & 3.1 & 2.6 \\
    RDI & 66.9 & \underline{98.4} & 98.7 & 83.1 & 35.6 & 42.5 & 25.4 & 0.8 & 22.4 & 3.2 & 9.8 & 10.8 \\
    RDI-SI & 70.1 & 98.0 & 98.9 & 89.3 & 53.3 & 57.3 & 40.0 & 4.5 & 40.7 & 8.8 & 21.4 & 27.4 \\
    RDI-VT & 66.4 & \underline{98.4} & 98.8 & 83.1 & 51.6 & 56.6 & 40.7 & 5.0 & 39.6 & 8.4 & 21.6 & 22.7 \\
    RDI-Admix & 72.9 & \underline{98.4} & \underline{99.1} & 88.0 & 46.3 & 58.3 & 36.5 & 1.8 & 33.4 & 4.5 & 15.4 & 17.1 \\
    RDI-ODI & 70.9 & 94.0 & 96.1 & 83.4 & 65.4 & 69.9 & 57.6 & 9.1 & 52.9 & 18.6 & 28.6 & 42.8 \\
    RDI-CFM & 86.1 & 98.3 & 97.0 & 92.1 & 71.7 & 79.9 & 72.0 & 10.3 & 66.5 & 20.3 & 41.9 & 45.8 \\
    \cmidrule(l){1-13}
    \rowcolor{Gray} RDI-FTM & \underline{87.5} & 97.8 & 96.7 & \underline{92.4} & \underline{75.4} & \underline{81.7} & \underline{74.0} & \underline{14.5} & \underline{71.0} & \underline{25.2} & \underline{49.3} & \underline{54.4} \\
    \rowcolor{Gray} RDI-FTM-E & \textbf{88.2} & 98.1 & 97.1 & \textbf{93.4} & \textbf{76.7} & \textbf{83.4} & \textbf{77.1} & \textbf{18.4} & \textbf{74.9} & \textbf{30.1} & \textbf{53.4} & \textbf{58.0} \\

    \toprule[0.15em]
    \addlinespace[0.4cm]
    \toprule[0.15em]
     \multirow{2}{*}{Attack}  & \multicolumn{12}{c}{RN-50 + LeViT $\Rightarrow$} &  \\\cmidrule(l){2-13}
                & VGG-16 & RN-50* & Inc-v3 & DN-121 & IR-v2 & Inc-v4 & Xcep & ViT & LeViT* & ConViT & Twins & PiT \\
    \midrule
    DI        & 71.3 & \textbf{99.2} & 29.5 & 80.3 & 16.2 & 23.7 & 15.4 & 1.2 & \textbf{100} & 6.7 & 18.2 & 20.8 \\   
    RDI       & 75.3 & 98.7 & 55.7 & 87.8 & 35.1 & 41.9 & 29.6 & 4.2 & 99.4 & 14.6 & 34.3 & 41.1 \\ 
    RDI-SI    & 78.1 & \underline{98.9} & 74.0 & 92.8 & 53.1 & 57.4 & 45.5 & 10.6 & \underline{99.9} & 30.5 & 52.2 & 67.1 \\
    RDI-VT    & 77.4 & \underline{98.9} & 66.4 & 88.1 & 50.7 & 56.0 & 48.4 & 14.6 & 99.1 & 30.3 & 53.9 & 60.7 \\
    RDI-Admix & 82.3 & 98.8 & 64.8 & 91.2 & 43.6 & 53.9 & 41.0 & 7.6 & \underline{99.9} & 17.4 & 43.9 & 48.8 \\ 
    RDI-ODI   & 81.9 & 96.6 & 81.6 & 89.6 & 69.9 & 73.4 & 67.9 & 27.8 & 97.4 & 52.7 & 66.5 & 81.0 \\
    RDI-CFM   & 88.5 & 98.8 & 84.1 & 92.1 & 69.5 & 78.7 & 72.1 & 26.4 & 99.8 & 51.6 & 77.5 & 81.3 \\
    
    \cmidrule(l){1-13}
    \rowcolor{Gray} RDI-FTM     &  \underline{89.2} & 98.4 & \underline{86.4} & \underline{93.0} & \underline{74.2} & \underline{81.0} & \underline{77.1} & \underline{41.0} & 99.5 & \underline{66.1} & \underline{85.5} & \underline{89.2} \\
    \rowcolor{Gray} RDI-FTM-E   &  \textbf{90.0} & 98.1 & \textbf{86.7} & \textbf{93.9} & \textbf{76.7} & \textbf{82.4} & \textbf{80.1} & \textbf{51.5} & 99.7 & \textbf{72.6} & \textbf{87.9} & \textbf{90.8} \\
    \bottomrule[0.15em]
    \end{tabular}%
    }
    \caption{
    Targeted attack success rates (\%) using ensemble-based surrogate models.
    All methods are combined with MI-TI.
    The best results are shown in bold, and the second best results are underlined.
    The surrogate models are marked with * in the first column.
    }
    \label{tab:ensemble}
    \vspace{-0.3cm}
\end{table*}

\clearpage

\begin{table*}[!t]
    \centering
    \resizebox{0.97\textwidth}{!}{%
    \begin{tabular}{clc|ccccccccccc}
    \toprule[0.15em]
    Source & Attack & Time (s) & VGG-16 & RN-18 & RN-50 & DN-121 & Xcep & MB-v2 & EF-B0 & IR-v2 & Inc-v3 & Inc-v4 & \textbf{Avg.} \\ \midrule
    \multirow{9}{*}{ViT} 
    & RDI        & 1.76 & 2.4 & 2.4 & 3.1 & 5.2 & 3.5 & 3.6 & 10.6 & 4.4 & 5.3 & 4.3 & 4.5 \\
    & RDI-VT     & 10.20 & 2.5 & 3.6 & 4.1 & 7.1 & 5.9 & 4.1 & 13.4 & 7.4 & 5.5 & 5.4 & 5.9 \\
    & RDI-Admix     & 5.11 & 5.4 & 5.4 & 6.9 & 11.5 & 6.7 & 6.9 & 18.7 & 8.1 & 10.1 & 8.0 & 8.8 \\
    & RDI-SI        & 8.38 & 4.2 & 8.2 & 9.5 & 17.6 & 10.2 & 9.1 & 24.7 & 13.3 & 18.0 & 12.1 & 12.7 \\
    & RDI-ODI        & 4.53 & 10.2 & 12.7 & 14.3 & 22.1 & 18.6 & 12.1 & 32.2 & \underline{22.5} & 22.3 & 20.7 & 18.8 \\
    & RDI-CFM    & 1.80 & 10.8 & 14.2 & 14.9 & 21.0 & 14.9 & 15.0 & 31.9 & 15.0 & 18.9 & 17.1 & 17.4 \\
    \cmidrule(l){2-14}
    & RDI-FTM                & 1.92  &  11.6 & 15.1 & 14.6 & 22.1 & 16.7 & 18.6 & 31.8 & 15.8 & 20.5 & 18.0 & 18.5 \\
    & RDI-FTM-E              & 3.81  &  \underline{13.4} & \underline{18.1} & \underline{18.3} & \underline{26.5} & \underline{19.6} & \underline{19.5} & \underline{37.2} & 19.8 & \underline{23.0} & \underline{21.3} & \underline{21.7} \\
    & RDI-FTM-E-SI$_{m_1=2}$  & 3.82  &  \textbf{24.0} & \textbf{29.1} & \textbf{30.0} & \textbf{41.5} & \textbf{28.9} & \textbf{31.3} & \textbf{54.7} & \textbf{30.9} & \textbf{39.3} & \textbf{31.8} & \textbf{34.2} \\
    \bottomrule[0.15em]
    \end{tabular}%
    }
    \caption{Targeted attack success rates (\%) against ten target models, with ViT as the surrogate model.
    All methods are combined with MI-TI.
    The best results are shown in bold, and the second best results are underlined.
    Time (s) denotes the average computation time required to attack a single image.
    RDI-FTM-E-SI$_{m_1=2}$ denotes using scaled input for RDI-FTM-E .
    }
    \label{tab:vit-cnn}
    \vspace{-0.3cm}
\end{table*}

\begin{table*}[!t]
    \centering
    \resizebox{0.57\textwidth}{!}{%
    \begin{tabular}{clc|cccccc}
     \toprule[0.15em]
    Source & Attack & Time (s) & ViT & LeViT & ConViT & Twins & PiT & \textbf{Avg.} \\
    \midrule
    \multirow{9}{*}{ViT}
    & RDI       & 1.76 & \textbf{99.5} & 24.3 & 23.3 & 11.7 & 26.0 & 37.0 \\  
    & RDI-VT    & 10.20 & 98.3 & 32.0 & 29.7 & 15.3 & 35.7 & 42.2 \\
    & RDI-Admix & 5.11 & \underline{99.1} & 39.0 & 34.6 & 18.5 & 40.5 & 46.3 \\
    & RDI-SI    & 8.38 & 98.4 & \underline{58.4} & \underline{69.2} & \underline{38.8} & \underline{63.8} & \underline{65.7} \\
    & RDI-ODI   & 4.53 & 87.6 & 45.9 & 42.1 & 29.5 & 49.3 & 50.9 \\
    & RDI-CFM   & 1.80 & 96.5 & 47.2 & 50.1 & 27.4 & 48.8 & 54.0 \\
    \cmidrule(l){2-9}
    & RDI-FTM                & 1.92 & 94.3 & 50.5 & 53.3 & 31.0 & 52.6 & 56.3 \\
    & RDI-FTM-E              & 3.81 & 96.0 & 54.9 & 55.6 & 36.3 & 55.8 & 59.7 \\
    & RDI-FTM-E-SI$_{m_1=2}$ & 3.82 & 96.3 & \textbf{71.2} & \textbf{74.8} & \textbf{54.7} & \textbf{74.6} & \textbf{74.3} \\

    \bottomrule[0.15em]
    \end{tabular}
    
    }
    \caption{Targeted attack success rates (\%) against five transformer-based DNNs, with ViT as the surrogate model. 
    All methods are combined with MI-TI.
    The best results are shown in bold, and the second best are underlined.
    Time (s) denotes the average time required to attack a single image.
    RDI-FTM-E-SI$_{m_1=2}$ denotes using scaled input for RDI-FTM-E .
    }  \vspace{-0.3cm}
    \label{tab:vit-vit}
\end{table*}

\clearpage

 \clearpage

\end{document}